\documentclass{article}

\usepackage[nonatbib,preprint]{neurips_2021}

\usepackage{times}
\usepackage{epsfig}
\usepackage{graphicx}
\usepackage{amsmath}
\usepackage{amssymb}
\usepackage[english]{babel}

\usepackage[utf8]{inputenc} 
\usepackage[T1]{fontenc}    
\usepackage{hyperref}       
\usepackage{url}            
\usepackage{booktabs}       
\usepackage{amsfonts}       
\usepackage{nicefrac}       
\usepackage{microtype}      

\usepackage{caption}
\usepackage{floatrow}
\usepackage{amsmath}
\usepackage{mathtools}
\usepackage{textcomp}
\usepackage{wrapfig}
\usepackage{graphicx}
\graphicspath{{figures/}}
\usepackage[ruled]{algorithm2e} 

\SetAlFnt{\small}
\SetAlCapFnt{\small}
\SetAlCapNameFnt{\small}
\SetAlCapHSkip{0pt}
\IncMargin{-\parindent}

\usepackage{algpseudocode}
\usepackage{multirow}
\usepackage{enumitem}
\usepackage{color}

\usepackage{amsmath,mathtools,amssymb,gensymb}
\newcommand{\norm}[1]{\left\lVert#1\right\rVert}
\newcommand*\rot{\rotatebox{90}}

\title{DV-Det: Efficient 3D Point Cloud Object Detection \\ with Dynamic Voxelization}

\author{%
  Zhaoyu Su \quad Pin Siang Tan \quad Yu-Hsing Wang \\
  DESR Laboratory \\
  Department of Civil and Environmental Engineering\\
  Hong Kong University of Science and Technology\\
\  \texttt{\{zsuad, pstan\}@connect.ust.hk; ceyhwang@ust.hk} \\
}

\begin{document}

\maketitle

\begin{abstract}
  In this work, we propose a novel two-stage framework for the efficient 3D point cloud object detection. Instead of transforming point clouds into 2D bird eye view projections, we parse the raw point cloud data directly in the 3D space yet achieve impressive efficiency and accuracy. To achieve this goal, we propose dynamic voxelization, a method that voxellizes points at local scale on-the-fly. By doing so, we preserve the point cloud geometry with 3D voxels, and therefore waive the dependence on expensive MLPs to learn from point coordinates. On the other hand, we inherently still follow the same processing pattern as point-wise methods (e.g., PointNet) and no longer suffer from the quantization issue like conventional convolutions. For further speed optimization, we propose the grid-based downsampling and voxelization method, and provide different CUDA implementations to accommodate to the discrepant requirements during training and inference phases. We highlight our efficiency on KITTI 3D object detection dataset with $75$ FPS and on Waymo Open dataset with $25$ FPS inference speed with satisfactory accuracy.
\end{abstract}

\section{Introduction}

Algorithms for point cloud detection are receiving increasing attention nowadays thanks to the blooming of robotics and the autonomous driving industry. Different from the detection algorithms on 2D images, detection on 3D point cloud is a non-trivial task due to the high sparsity and unstructured property of point clouds. Meanwhile, detection algorithms are expected to run at the real-time speed, while they are usually deployed on the edge devices with limited computation resources, which induces exigent demand for the computation efficiency. 

\begin{figure}
  \centering
  \includegraphics[width=\textwidth]{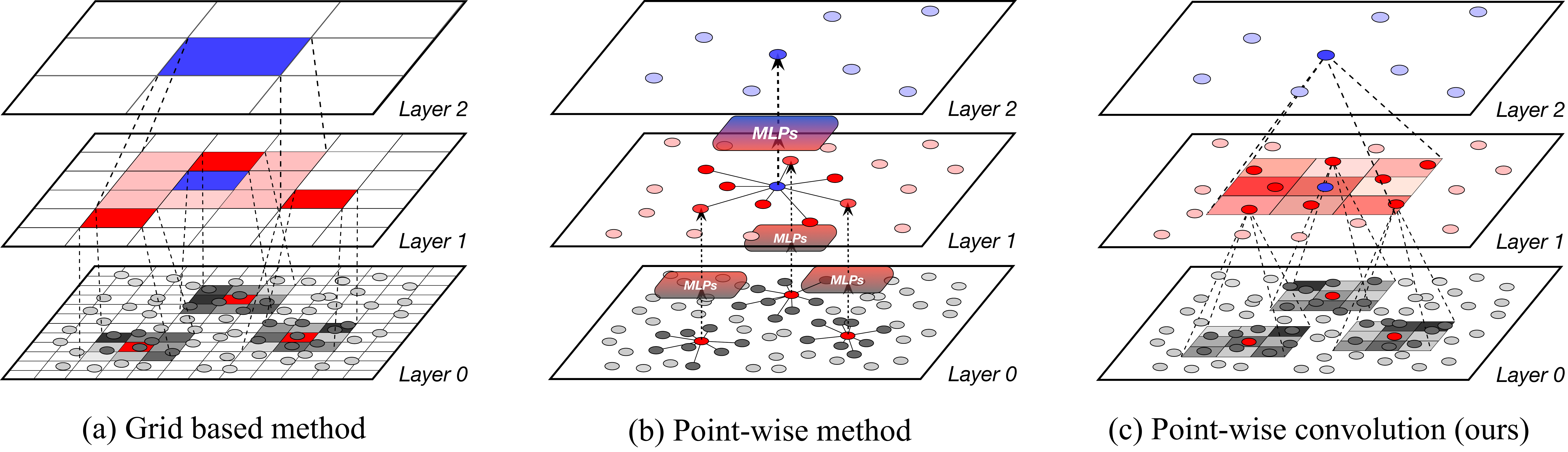}
  \caption{Comparison between different sampling strategies and network architectures. (a) Global voxelization for most of the grid based methods, notice the severe quantization issue at layer 2. (b) Point-wise method, commonly used for the point-wise MLPs based methods like PointNet; (c) Our proposed point-wise convolution based on the dynamic voxelization operation.}
  \label{fig:compare}
\end{figure}

The existing point cloud detection methods can be roughly classified into two groups: grid based methods and point-wise methods. Grid based methods basically follow the same pattern as the 2D image cases: they transform the point cloud into regular grid representations like 3D voxels\cite{Chen2017, Shi2019a, zhou2018voxelnet, yan2018second} or 2D bird eye view (BEV) projections\cite{Lang2019, yang2018pixor, yang2018hdnet}, and then use 3D or 2D convolutional neural networks (CNNs) to extract the features and obtain bounding box predictions. However, the transformation from the point cloud to regular grids leads to the information loss, and due to the quantization issue, the feature maps at high-level CNN layers often lack the precise regional feature representations (Fig.\ref{fig:compare} a), which is essential for an accurate bounding box regression. 

On the other hand, developed from the pioneering work PointNet\cite{Qi2017, Qi2017a}, more and more point-wise methods are proposed recently, which directly take raw 3D LiDAR points as input and learn point-wise feature representations from point coordinates via multi-layer perceptrons (MLPs)\cite{qi2018frustum, Shi2019b, yang2019std}. As shown in Fig.\ref{fig:compare} b, point-wise methods no longer suffer from the quantization problem as conventional convolutions, but this comes at a price: learning representations from unstructured point coordinates is a hard task, and therefore, heavy MLPs architectures with expensive computational cost are usually involved in these methods, leading to relatively slow inference speed.

There have been attempts that try to combine grid based methods and point-wise methods, e.g., in PV-RCNN\cite{shi2020pv} , these two methods are leveraged in an alternate pattern at the \textit{framework level} to achieve the best performance. Despite the impressive accuracy, heavy MPLs still remain in \cite{shi2020pv}, while the inference speed is still limited. Currently, the mainstream point cloud detection methods run at a speed of ~20 FPS for the KITTI dataset (few can reach ~45 FPS\footnote{Without TensorRT optimization.}\cite{Lang2019}) on a powerful dedicated GPU, with $90\degree$ horizontal front-view FOV. This is apparently not sufficient for the full $360\degree$ FOV scenario with a LiDAR sampling rate of 20Hz in reality, not to mention the limited computation hardware resources on the edge devices. 

In this work, we propose a new method that combines the advantages from both grid based methods and point-wise methods at the \textit{perceptron level}. As shown in Fig.\ref{fig:compare} c, given a point cloud, we first extract key-points and gather neighbouring points around each key-point just like the point-wise methods. However, instead of learning from point coordinates with MLPs, we leverage local-scale 3D voxels to preserve the 3D geometry and then learn features with 3D convolutions. By doing so, we waive the requirements for the expensive MLPs and tackle the quantization issue as well. 

Just like other point-wise methods, we also face the same problem: how to extract key-points from hundreds of thousands of input points efficiently. In addition, we also need to figure out how to construct 3D voxels on-the-fly during the network propagation at real-time. To this end, we propose the \textit{grid based downsampling} and hierarchical point-wise convolution with \textit{dynamic voxelization}, to take the place of commonly adopted farthest point sampling (FPS) + k-nearest-neighbours (k-NN) combination, which is time consuming and memory-hungry. In addition, as a two-stage detection framework, we also develop location-aware RoI pooling, a light-weight pooling method that is 3$\times$ faster and 4$\times$ more memory efficient than the previous work without losing accuracy significantly. 

Moreover, for further performance improvement, we adopt 3D IoU loss\cite{zhou2019iou} for bounding box regression. We develop an efficient algorithm for 3D IoU loss calculation, which is fully based on the native TensorFlow operations and can be implemented with less than 200 lines of Python code. The tedious back propagation calculation of 3D IoU can just be left to the deep learning frameworks and no longer has to be manually implemented. 

Powered by all the innovations above, we propose DV-Det, a two-stage detection framework for efficient 3D object detection from point clouds. Our method achieves the impressive 75 FPS inference speed on the KITTI dataset\cite{Geiger2012} and 25 FPS on the Waymo Open Dataset \cite{Sun2020}, while still maintains satisfactory accuracy performance.

\section{Related Works}
\subsection{Point-wise Based 3D Detection}
The point-wise 3D detection methods has gained a lot of attention after the success of PointNet \cite{Qi2017}, and PointNet is still the most widely adopted perceptron unit in many point-wise detection methods. Frustum-PointNet\cite{qi2018frustum} is the first work to apply PointNet into 3D detection, it crops 3D frustums based on the detection results from 2D images, and then use PointNet to parse the frustums point cloud data. STD\cite{yang2019std} employs PointNet++\cite{Qi2017a} as the backbone for features extraction, and uses PointsPool for efficient proposal refinement. Similarly, \cite{Shi2019b, zhou2018voxelnet} also leverages PointNet++ to generate RoI proposals directly from the input point clouds, while \cite{qi2019deep} uses a voting strategy for better feature aggregation returned from PointNet++. 

However, PointNet (including its variants) relies on the MLPs (functioning like a sub-network in each layer) to learn the geometrical features from the point coordinates, and it is therefore a computationally intensive method. Hence, the point-wise methods mentioned above usually run only at $\sim$10 FPS on the KITTI dataset with a dedicated GPU.

\subsection{Grid Based 3D Detection}
Convolutions are particularly good at dealing with data organised in regular grids. By converting the unstructured point cloud into 3D voxels, the geometry of points is preserved within the arrangement of voxels, and therefore, it no longer has to be learnt from point coordinates like PointNet. In previous works\cite{Shi2019a, yan2018second, He2020}, 3D point clouds are converted into 3D voxel representations, and 3D sparse convolutions\cite{Graham2017} are used as backbones for feature extractions. 

By converting the 3D points into 2D bird eye view (BEV) projections\cite{Lang2019, yang2018hdnet, yang2018pixor}, most of the methods originally proposed for detection on 2D images can be transferred to the 3D seamlessly. However, converting the entire point cloud into 2D projections means the permanent loss of 3D geometry information, and therefore, the existing 2D BEV based methods usually show relatively low accuracy on the benchmark datasets, especially for the hard cases only with few points. There are also works that adopt similar implementation, and fuse the LiDAR point cloud with 2D camera images at the feature map level\cite{Liang2019, Liang2018deep}, but the interpretation of high-resolution camera images is a difficult task, which usually leads to unsatisfactory inference speed.

\subsection{Rethink About Existing Methods}
Point-wise methods are able to learn precise point-wise feature representations, however, their dependencies on MLPs often lead to heavy network architectures. Meanwhile, expensive operations like k-NN and FPS are also involved for point downsampling and feature aggregation, which further drags down the inference efficiency.

On the other hand, the grid based methods are more efficient, but due to the feature dilution issue, feature maps at high-level network layers often possess low resolution, which obstructs the network learning precise local feautre representations. E.g., in work \cite{Shi2019a}, although the RoI proposals have already been predicted at the bottle-neck of its U-Net \cite{Ronneberger2015} style network architecture, but the feature maps are still upsampled to provide accurate enough information for the RoI-pooling in second stage.

In this work, we propose a novel method which maintains precise feature representations like point-wise methods without the requirement for complicated MLPs, while avoids the general issues existing in the grid base methods at the same time.

\section{Proposed Method}
Learning from point coordinates with MLPs is too expensive for a rapid detection framework to adopt, therefore, we stick to the grid based convolution approach as our basic perceptron unit. However, instead of converting the entire point cloud into 3D voxels, we adopt similar strategy as point-wise methods: we only construct local-scale 3D voxels around the key-points of the point cloud, and parse 3D features with dense 3D convolution.  In this way, we fuse the point-wise methods and grid based methods at the perceptron level, and achieve remarkable efficiency improvement. 

Nevertheless, just like other point-wise methods, we are facing the same problem: how to select key-points from a point cloud with hundreds of thousands of points. Moreover, we also need to construct the 3D convolution kernels on-the-fly during the forward propagation efficiently enough to achieve the real-time performance. In this section, we will give our answers to these two questions in section \ref{sec:downsampling} and \ref{sec:voxelization} respectively, followed by the illustrations to our Location-Aware RoI pooling in section \ref{sec:roi-pooling} and 3D IoU loss function in section \ref{sec:iou-loss}. 

\begin{figure}
  \centering
  \includegraphics[width=\textwidth]{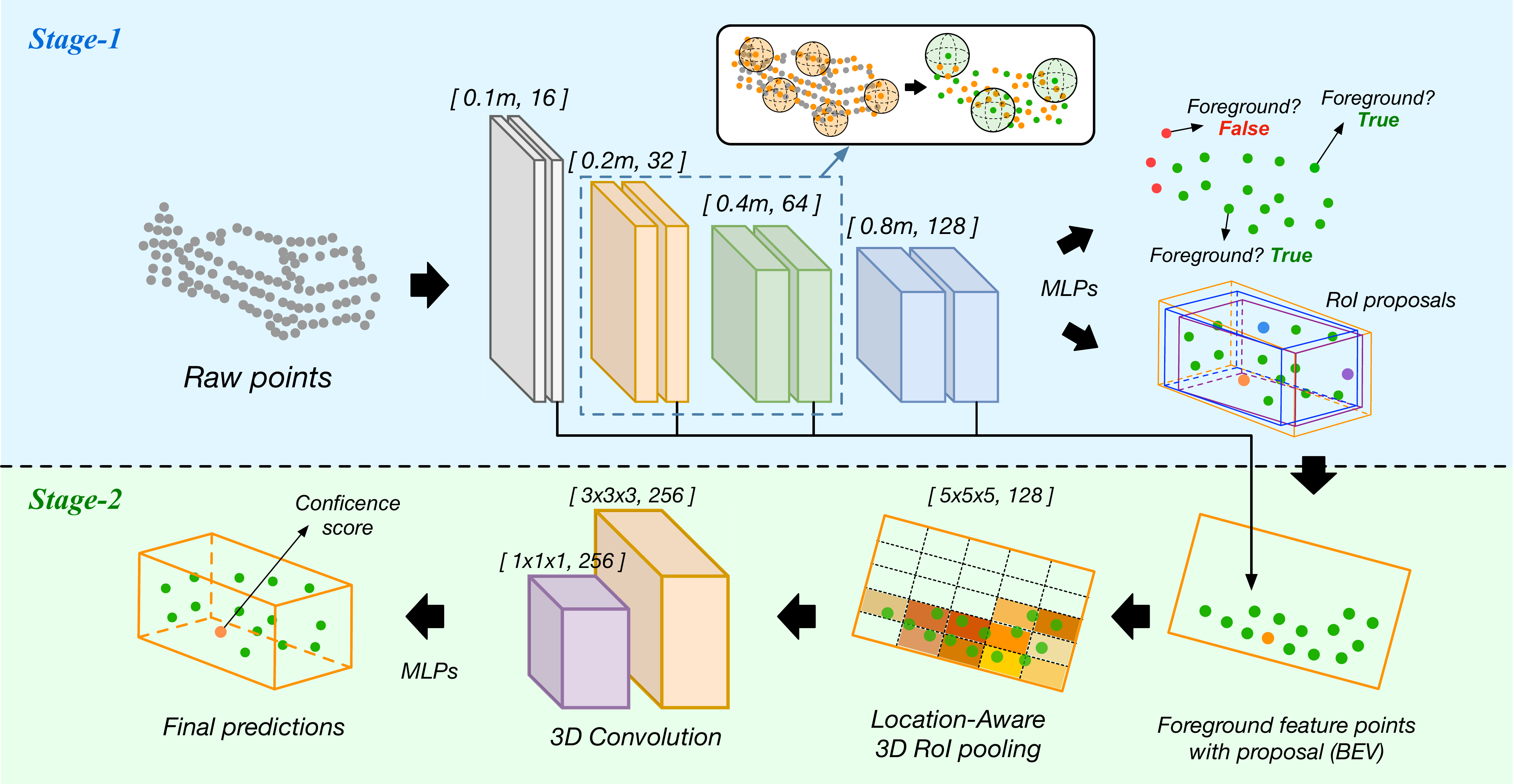}
  \caption{Visualization of the proposed two-stage DV-Det framework. In the first stage, raw 3D points are processed by point-wise convolutional layers with dynamic voxelization to extract features and obtain 3D bounding-box proposals. In the second stage, the proposals are refined based on the Location-Aware RoI pooling method to get the final prediction results.}
  \label{fig:architecture}
\end{figure}

\subsection{Grid Based Point Downsampling}
\label{sec:downsampling}
Farthest point sampling (FPS) is the most widely adopted algorithm for point cloud downsampling\cite{Qi2017}, however, as an algorithm with complexity $O(n^2)$, it runs very slowly for point cloud at large scales, making it a straggler in the real-time detection framework. On the other hand, random downsampling is efficient enough with $O(n)$ complexity, but it is sensitive to the local point density and often leads to unstable performance\cite{hu2020randla}. In this work, we propose the grid based point downsampling method, which runs almost as efficiently as random sampling, while keeps similar functionalities as FPS.

\begin{figure}
  \centering
  \includegraphics[width=\textwidth]{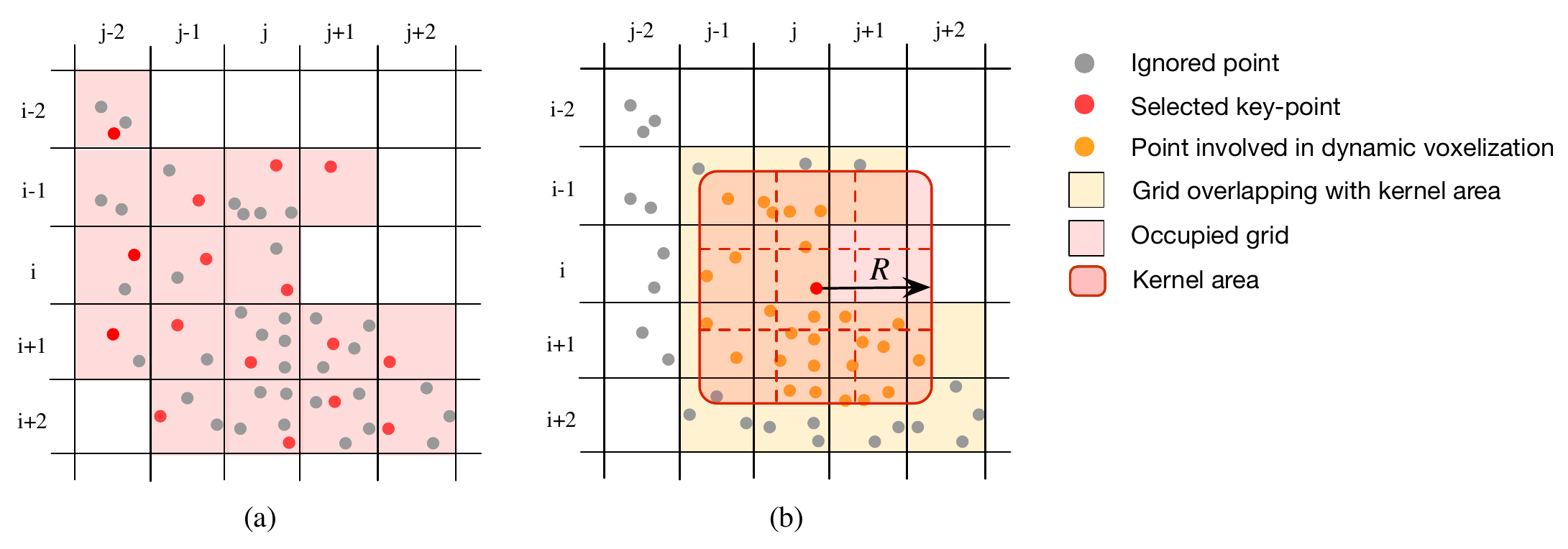}
  \caption{Visualization of grid based point downsampling (a) and dynamic voxelization (b).}
  \label{fig:voxelization}
\end{figure}

Specifically, for a point cloud $S=\{(p_i, f_i):i=1,2,...,N\}$, where $p_i\in \mathbb{R}^3$ is the point coordinate, $f_i\in \mathbb{R}^c$ is the point-wise feature with channel $c$ and $N$ is the total number of points, given the downsampling resolution $r$, we first divide the entire point cloud into 3D regular grids with grid length $r$ as shown in Fig.\ref{fig:voxelization} a.

For every point $p_i$ in ${P}$ with coordinate $[p_i^x, p_i^y, p_i^z]$, we obtain its corresponding grid index as $[\lfloor\frac{p_i^x}{r}\rfloor, \lfloor\frac{p_i^y}{r}\rfloor, \lfloor\frac{p_i^z}{r}\rfloor]$, which indicates the grid that point will fall into. Next, we randomly select one point in each grid as our downsampling output, unless the grid is not occupied by any points. This strategy may sound straightforward, but not the implementation, especially for the balance between the time and space complexity. Therefore, we adopt different implementations for inference and training phase separately due to their different needs for speed and memory efficiency.

For the inference purpose, speed matters more than memory efficiency, and therefore, our implementation creates a 3D grid buffer with size $\lfloor\frac{W}{r}\rfloor \times \lfloor\frac{L}{r}\rfloor \times \lfloor\frac{H}{r}\rfloor$ first, where the $W$, $L$ and $H$ stand for the dimensions of the input point cloud; then, we calculate the grid index for every point which are fitted into the corresponding grid buffer location. Each grid buffer location is writable for only once, which means its value cannot be modified once it has been occupied, so that the other points with the same grid index will not overwrite the same buffer again. With this setup, we achieve the $O(n)$ complexity, but the speed also comes at a price: the 3D grid buffer has to be pre-allocated in the memory space first. Fortunately, it usually will not be a problem for most of the cases, e.g., for Waymo Open Dataset with input range of 150m $\times$ 150m $\times$ 6m, when the $r$ is 0.1m, the total RAM consumption by our inference implementation is around 500MB (with float32), while the speed is 5$\times$ faster than our training implementation (introduced below). Another good news is that we can reuse the same downsampling results in the subsequent model layers, as long as the downsampling ratio $r$ remains unchanged. 

By contrast, in the training implementation, we aim at fitting as large batch size as possible into GPU, while the time consumption is less important. Following this principle, we calculate the grid index $[\lfloor\frac{p_i^x}{r}\rfloor , \lfloor\frac{p_i^y}{r}\rfloor ,\lfloor\frac{p_i^z}{r}\rfloor]$ for every point first, then sort the points according to their grid indexes, and collect the points with unique grid indexes as the final downsampling results. By doing so, we waive the requirement for the memory-hungry grid buffer implementation. Since the sorting operation is involved here, the approximate complexity is $O(nlogn)$, not as ideal as $O(n)$, but still much faster than the FPS. 

\subsection{Hierarchical Convolution with Dynamic Voxelization}
\label{sec:voxelization}
The purpose of our hierarchical point-wise convolution is to combine the advantages of both conventional convolutions and the point-wise operations. Instead of applying these two techniques in an alternate way like PV-RCNN\cite{shi2020pv}, in this work, we directly deploy 3D convolutions in a point-wise fashion. Specifically, after the extraction of key-points via downsampling, we first construct local-scale 3D voxels around each key-point with our proposed dynamic voxelization operation, then we parse the point-wise 3D voxel data with dense 3D convolution to obtain the feature representations. 

\subsubsection{Dynamic Voxelization}
Given the input point cloud $S_1=\{(p_1^i, f_1^i):i=1,2,...,N_1\}$, $p_1^i\in \mathbb{R}^3$, $f_1^i\in \mathbb{R}^{c_1}$, with the subscript "1" suggesting it is the first layer in the network, we first downsample $\{p_1^i\}$ using our grid-based downsampling with the downsampling resolution $r_1$, and get the downsampled key point set $\{p_2^i:i=1,2,...,N_2, N_2\leq N_1\} \subseteq \{p_1^i:i=1,2,...,N_1\}$. Next, for every $p_2^i$, we query its neighbouring points in $S_1$ within a fixed radius $R_1$: $S_1^i=\{(p_1^j, f_1^j):j=1,2,...,N_1^i | \norm{p_1^j-p_2^i}\leq R_1\}$, and then fit $S_1^i$ into the 3D regular voxels with resolution $k\times k\times k$. By doing so, we convert the original $S_1$ into a set of point-wise 3D voxels: $V_1=\{(p_2^i, v_1^i),i=1,2,...,N_2\}$, where $v_1^i\in \mathbb{R}^{k\times k\times k\times c_1}$  encodes the local point cloud geometry around the key point $p_2^i$. 

The key of our dynamic voxelization lies on how to construct the 3D voxels $v^i$ for each key-point $p^i$. However, even after the grid based downsampling operation, the total number of remaining key-points is still considerable ($\sim$ 100k at most), and to perform local-scale voxelization with so many key-points in real-time, we propose the grid based voxelization operation.

As suggested by its name, grid based voxelization basically shares a similar implementation as our grid based downsampling operation. As shown in Fig.\ref{fig:voxelization} b, given the kernel radius $R$ and kernel resolution $k$, we divide the point cloud into regular grids with resolution $2R/k$ (defined as kernel resolution in this context), and for any given kernel centroid and query radius $R$, we fit interior points into the  $k \times k \times k$ 3D voxels. In regard to the voxels occupied by multiple points, the average-pooling strategy is used. 

To accelerate this process, we only need to go through the points that lies in the grids which overlap with the kernel area, meanwhile, we also apply different implementations separately to fulfill the different preferences during training and inference phases, and readers may refer to the source code for more details. 

A similar idea to our dynamic convolution is also introduced in \cite{hua2018pointwise}, however, the voxelization is only performed once for all the layers before the network propagation in \cite{hua2018pointwise}, therefore, all the layers in the network always share the same receptive field and there is no concept of hierarchy. By contrast, in our work, the voxelization is performed on-the-fly in every convolutional layer with highly efficient implementations, and therefore the receptive field dynamically changes along the model depth.

\subsubsection{Point-wise 3D Convolution}
As for the convolution part, taking the first convolutional layer as an example, we convolve the point-wise 3D voxel $v_1^i$ with dense 3D kernel $W_1 \in \mathbb{R}^{k\times k\times k\times c_1 \times c_2}$.  $W_1$ has the same $k \times k \times k$ resolution as $v_1^i$, while $c_1$ and $c_2$ are the input and output feature channels respectively. After convolution, we have   $f_2^i=v_1^i*W_1$, and  $f_2^i \in \mathbb{R}^{c_2}$. Finally, we obtain a new point set  $S_2=\{(p_2^i, f_2^i):i=1,2,...,N_2\}$, which can be treated as the input of another successive convolutional layer. Since the convolution output features are closely associated with the point coordinates, the output $f^i$  now exactly represents the features at location $p_i$ , instead of a large pooling area, like conventional convolutions in Fig.\ref{fig:compare} a. A hierarchical arrangement of point-wise convolutional layers shapes our backbone and region proposal network (RPN) as shown in the Fig.\ref{fig:architecture}.

\subsection{Location-Aware RoI Pooling}
\label{sec:roi-pooling}
As a two-stage detection framework, RoI pooling plays an essential part in the second refinement stage. However, to perform efficient yet accurate RoI pooling still remains an open problem for 3D point cloud detection. The concept of RoI pooling was officially introduced in \cite{Shi2019a}, and later on was adopted in the successive works\cite{shi2020pv}. 

\begin{wrapfigure}{R}{0.5\linewidth}
\includegraphics[width=\textwidth]{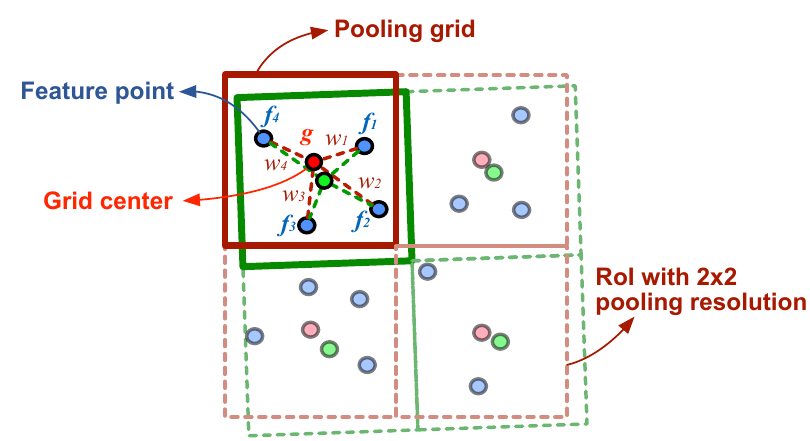}
{\caption{Location-Aware RoI pooling illustrated in 2D with two proposals and pooling resolution of 2$\times$2. With LA-RoI pooling, our network can capture the tiny variance between to the two proposals that are very close to each other.}\label{fig:roi-pooling}}
\end{wrapfigure}

 But how to efficiently aggregate the point-wise features into RoI pooling grids still remains a problem. In \cite{shi2020pv}, features in each pooling grid are learnt from additional convolutions, e.g., for RoI pooling with grid resolution of $6\times 6\times 6$, a total number of 216 convolution calculations are involved, and that is for one single RoI only, therefore, it is unaffordable to an efficient detection framework. 

To this end, we propose the Location-Aware RoI Pooling method. As shown in Fig.\ref{fig:roi-pooling}, during the pooling operation, we assign additional weight to the input point-wise features according to their Euclidean distances to the pooling grid centre. Specifically, given a 3D RoI of dimension $W\times L\times H$ and the target pooling grid resolution $k$, we first gather all the interior point-wise features $\{f_i\}$ within a certain pooling grid $g$ among the $k\times k\times k$ grids; next, we calculate representative features $f_g$ as: 
\begin{equation}
f_g=\frac{1}{n}\sum_{i=1}^{n}{w_if_i};w_i=e^{1-\frac{d_i}{r}};r=max\{\frac{W}{k}, \frac{L}{k}, \frac{H}{k}\} \label{eq:la_pooling}
\end{equation}
where $d_i$ is the Euclidean distance between the feature point $f_i$ and the grid centre $g$, $r$ is the maximum dimension of pooling grid among $W$, $L$ and $H$ orientations. We normalize $f_g$ by the total number of interior points $n$ in each pooling grid, so that the scale of $f_g$ will not be heavily affected by $n$. In practice, the largest value of $n$ is set as 5, which means for each grid $g$, only at most 5 interior point-wise features $f_i$ are taken into consideration, and the redundant points are simply omitted. By doing so, we are able to capture the tiny location perturbations between different RoIs, while avoid the expensive convolution computations. 

\begin{algorithm}[h!]
	\SetKwInOut{Input}{Input}
	\SetKwInOut{Output}{Output}
	\Input{Predicted bounding box $B^p$ and the corresponding ground truth $B_g$: \\ $B^p=(x^p,y^p,z^p,w^p,l^p,h^p,r^p)$, $B^g=(x^g,y^g,z^g,w^g,l^g,h^g,r^g)$}
	\Output{The 3D IoU loss between $B^p$ and $B^g$: $L_{IoU}$} 
	\begin{algorithmic}[1]
		\State Convert $B_p$ and $B_g$ to BEV representation: $B^p_{bev}=(x^p,y^p,w^p,l^p,r^p)$, $B^g_{bev}=(x^g,y^g,w^g,l^g,r^g)$
		\State Get the vertices of $B^p_{bev}$ and $B^g_{bev}$. For $i\in\{1,2,3,4\}$:\newline $\mathbf{V}_i^p=[(-1)^{i}w^p/2, (-1)^{i-1}l^p/2] + [x^p,y^p]$; $\mathbf{V}_i^g=[(-1)^{i}w^g/2, (-1)^{i-1}l^g/2]+[x^g,y^g]$
		\State Get the perspective coordinates of $\mathbf{V}^p$ regarding its relative position to $B^g_{bev}$ as $\mathbf{V}^{p\rightarrow{g}}$:\newline
		$\mathbf{V}^{p\rightarrow{g}}=\mathbf{R}_{(r^p-r^g)}[\mathbf{V}^p]^T+[x^p-x^g,y^p-y^g]$ (\textit{$\mathbf{R}_{*}$ is the rotation transformation regarding angle $*$})
		\State Get the edge extensions of $B^p_{bev}$ with vertices $\mathbf{V}^{p\rightarrow{g}}$. For $i\in\{1,2,3,4\}$: \newline
		$y=(-1)^{i}\frac{\mathbf{V}_i^{p\rightarrow{g}}[1]-\mathbf{V}_{i-1}^{p\rightarrow{g}}[1]}{{\mathbf{V}_i^{p\rightarrow{g}}[0]-\mathbf{V}_{i-1}^{p\rightarrow{g}}[0]}}x+(-1)^{i-1}\frac{\mathbf{V}_{i-1}^{p\rightarrow{g}}[1]\mathbf{V}_{i}^{p\rightarrow{g}}[0]-\mathbf{V}_{i}^{p\rightarrow{g}}[1]\mathbf{V}_{i-1}^{p\rightarrow{g}}[0]}{{\mathbf{V}_i^{p\rightarrow{g}}[0]-\mathbf{V}_{i-1}^{p\rightarrow{g}}[0]}}$
		\State Get all the intersections $\{\mathbf{I}_j^g\},j\in\{1,2,...,16\}$ between $B^p_{bev}$ and $B^g_{bev}$ by setting $x$ as $\pm \frac{1}{2}w^g$ or setting $y$ as $\pm \frac{1}{2}l^g$ in \textbf{Step 4} (\textit{the superscript $g$ indicates $\mathbf{I}^g$ is represented based on its relative position to $B^g_{bev}$})
		\State Concatenate $\mathbf{I}^g$, $\mathbf{V}^{p\rightarrow{g}}$ and $\mathbf{V}^g$ as $\mathbf{P}^g$=\{$\mathbf{I}^g$, $\mathbf{V}^{p\rightarrow{g}}$,$\mathbf{V}^g$\}, filter out the points in $\mathbf{P}^g$ that are outside $B^g_{bev}$, and that leads to $\mathbf{P}^g=\{\mathbf{P}_k^g|k\in\{1,2,...,24\}\cap \norm{\mathbf{P}_k^g[0]}\leq \frac{1}{2}w^g\cap \norm{\mathbf{P}_k^g[1]}\leq \frac{1}{2}l^g\}$
		\State Similar to \textbf{Step 3}, get the perspective coordinates of $\mathbf{P}^g$ regarding its relative position to $B^p_{bev}$ as $\mathbf{P}^{g\rightarrow{p}}$
		\State Repeat \textbf{Step 6} to filter out the points in $\mathbf{P}^{g\rightarrow{p}}$ that are outside $B^p_{bev}$, and that returns a new set $\mathbf{P} \subset{\mathbf{P}^{g\rightarrow{p}}}$, which are the vertices of the polygon intersection area between $B^p_{bev}$ and $B^g_{bev}$
		\State Sort the points in $\mathbf{P}$ counterclock-wisely and calculate the area of the intersection polygon using \textbf{Shoelace algorithm}: $A=\frac{1}{2}\left| (\sum\limits_{i=1}^{n-1}\mathbf{P}_i[0]\mathbf{P}_{i+1}[1])+\mathbf{P}_n[0]\mathbf{P}_{1}[1]- (\sum\limits_{i=1}^{n-1}\mathbf{P}_{i+1}[0]\mathbf{P}_i[1])-\mathbf{P}_1[0]\mathbf{P}_{n}[1]\right|$
		\State Get the intersection height $H$ between $B^p$ and $B^g$, then calculate the 3D IoU as:\newline $IoU_{3D}=\frac{A\times H}{w^p\times l^p\times h^p + w^g\times l^g\times h^g - A\times H}$, and $L_{IoU}=1-IoU_{3D}$

	\end{algorithmic}
	\caption{3D IoU Loss Calculation}
	\label{alg:iou_loss}
\end{algorithm}

\subsection{Loss Functions}
\label{sec:iou-loss}
As a two-stage model, we use multi-task loss functions to train our model in both stages. During the first stage, we try to differentiate foreground points from the point cloud and generate RoI proposals for every foreground points at the same time. As for loss function for the RoI regression optimization, we use the 3D IoU loss\cite{zhou2019iou} as our optimization target, instead of the commonly adopted smooth-L1 loss. 

\textbf{3D IoU Loss.}~
The IoU loss has been proven to be more suitable for detection tasks than smooth-L1 loss\cite{zhou2019iou, rezatofighi2019generalized}, as it naturally reflects the bias between a predicted bounding box and its corresponding ground truth. The calculation itself for 3D IoU is not difficult, but as a loss function, being differentiable is a prerequisite and it is not an easy task. In previous works\cite{zhou2019iou}, the back propagation of 3D IoU has to be implemented manually with C++/CUDA extensions, which can be tedious and fallible. To tackle this problem, here we propose an efficient algorithm for 3D IoU calculation, which is fully based on the native operations supported by the modern deep learning framework like TensorFlow and PyTorch. This means that we no long have to implement the back propagation manually and can safely leave this task to the auto differentiation functions of modern deep learning frameworks. 

Concretely, given a pair of predicted bounding boxes $B^p=(x^p,y^p,z^p,w^p,l^p,h^p,r^p)$ and the corresponding ground truth $B^g=(x^g,y^g,z^g,w^g,l^g,h^g,r^g)$, where $[x^*, y^*, z^*]$ is the geometry centroid of bounding box $*$; $w^*$, $l^*$ and $h^*$ are the 3D dimensions and $r^*$ is the rotation angle around $z(yaw)$ axis (assuming only have rotations along $z$ axis), we calculate the 3D IoU loss according to the algorithm \ref{alg:iou_loss}. The key of our algorithm lies on how to obtain the intersections efficiently between two bounding boxes on 2D BEV plane, and to achieve this goal, we transform the vertex coordinates of two bounding boxes based on their relative positions to each other via rotations and translations. By doing so, the intersections can be easily determined by setting restrictions on $x$, $y$ coordinates, and more importantly, our algorithm can be executed in batch-wise pattern without looping through every bounding box, which ensures its usability in applications. 

The multi-task loss function we use for the firs stage optimization is:
\begin{equation}
L_{stage1}=L_{cls} + \alpha L_{iou} + \beta L_{rot} \label{eq:stage1_loss}
\end{equation}
where $L_{cls}$ is the focal loss for foreground classification and $L_iou$ is the 3D IoU loss. $L_{rot}$ is the smooth-L1 loss for rotation angle regression and $L_{rot}=$\textit{smooth-L1}$(\sin{(r_p-r_g)})$, which prevents $L_{iou}$ mixing up the $w$ and $l$ dimension. As for the second stage, the loss function is defined as follow:
\begin{equation}
L_{stage2}=L_{conf} + \alpha L_{iou} + \beta L_{rot} + \gamma L_{flip} \label{eq:stage2_loss}
\end{equation}
where $L_{conf}$ is the loss for IoU guided confidence score [pvrcnn], while $L_{iou}$ and $L_{rot}$ follow the same definitions as before. $L_{flip}$ is the cross-entropy loss for binary classification, which decides whether the predicted bounding box shall be rotated by 180\degree. The reason why $L_{flip}$ is needed is that $L_{rot}$ only regulates on the $\sin$ value of $r_p-r_g$, and the network cannot differentiate the correct orientation of bounding boxes basing on $L_{rot}$ alone. In practice, we set $\alpha=2$, $\beta=\gamma=0.5$. $L_{cls}$ and $L_{conf}$ are calculated for all the output points, while $L_{iou}$, $L_{rot}$ and $L_{flip}$ are only calculated for the foreground points. 


\section{Experiments}
\subsection{Implementation Details}
\textbf{Network Architecture.}~
As shown in Fig.\ref{fig:architecture}, our backbone network is based on the hierarchical point-wise convolutional layers with dynamic voxelization. The backbone is composed of four convolution blocks, and each block includes two point-wise convolutional layers. The downsampling resolution $r$ is set as 0.1m, 0.2m, 0.4m, 0.8m for each block, and the kernel resolution $k$ is 3 for all the blocks with the number of feature channels 16, 32, 64 and 128. The output features from each block are then concatenated and fed into the second stage as input to the LA-RoI pooling layer. The LA-RoI pooling transforms the point-wise features into $5\times 5\times 5$ dense 3D voxels, and pass them to the dense $3\times 3\times 3$ 3D convolutions. We apply “valid” padding strategy for the 3D convolution and the voxel dimensions are reduced to $1\times 1\times 1$ after two successive dense convolutional layers. Finally, we feed these features into MLPs to return the final predictions. 

\textbf{Training and Inference Setup.}~
We train our DV-Det in an end-to-end fashion from scratch on the KITTI 3D object detection dataset\cite{Geiger2012} and Waymo Open dataset\cite{Sun2020}. The model is trained on 16 NVIDIA RTX 2080 Ti GPUs with 80 training epochs and the learning rate of 0.01. Batch size is set as 64 for KITTI dataset and 32 for Waymo dataset. In the second stage, we resample the negative and positive proposals with a ratio of 2:1 and IoU threshold of 0.5. During training phase, the input data are augmented with random rotations along $z$ axis between range $[-\frac{\pi}{4},\frac{\pi}{4}]$ and random scaling along three axes in range $[0.95, 1.05]$. Following the practice in \cite{shi2020pv, yan2018second}, we also randomly "paste" ground truth objects collected from different scenes (within the same dataset) to enrich the training data. We train our model with the "training" implementations of grid based downsampling and dynamic voxelization, and run inference with the "inference" implementations. 

\subsection{Evaluation on KITTI Dataset}
We first evaluate our performance on the KITTI 3D object detection dataset, which comprises two partitions: training dataset (7,481 samples) and testing dataset (7,518 samples). We further divide the training partition into two sets: \textit{train} set with 3,712 samples and \textit{val} set with 3,769 samples. In training phase, we utilize 80\% of the samples in the \textit{train}+\textit{val} combination as training data, and leave the rest 20\% for validation. We evaluate the performance based on the testing dataset from the KITTI official online evaluation server, and report our accuracy on \textit{car} and \textit{cyclist} class in Table \ref{tab:kitti}. 

We only compare with the pure point cloud based methods, and we divide the existing methods into two categories: "3D Perceptron" are the methods that parse the point cloud in 3D space using perceptrons like 3D convolution or PointNet; "BEV" are the methods that convert 3D point cloud onto 2D BEV plane, the process data with 2D convolutions. We notice that all of the existing 3D perceptron based methods have a inference speed less than 25 FPS on a dedicated GPU, and considering the 90\degree front-view FOV in KITTI dataset, we can hardly say they are ready for the real-world applications. On the other hand, BEV based methods run at considerably faster speed than the 3D perceptrons based methods, however, converting 3D point cloud into 2D BEV projections means the permanent loss of 3D spatial information, and that usually leads to unsatisfactory performance on hard examples. 

By contrast, our method achieves remarkable 75 FPS on the KITTI dataset, which is 3$\times$ faster than the state-of-the-art 3D perceptron based method. Meanwhile, compared with the existing BEV based methods, we overperform them regarding all the evaluation metrics and achieve 45\% faster inference speed than the widely adopted PointPillar method. 

\begin{table*}
	\small 
	\vspace{-2mm}
	\begin{center}
		\scalebox{0.77}[0.77]{
			\setlength\tabcolsep{5pt}
			\begin{tabular}{c|c|c|ccc|ccc|ccl|ccl}
				\hline
				&
				\multirow{2}{*}{Method} & 
				Speed &
				\multicolumn{3}{c|}{~~Car - 3D Detection ~~} & \multicolumn{3}{c|}{~Car - BEV Detection~} & \multicolumn{3}{c|}{~Cyclist - 3D Detection~} & \multicolumn{3}{c}{Cyclist - BEV Detection}\\
				&&(FPS)&Easy & Mod. & Hard & Easy & Mod. & Hard & Easy & Mod. & Hard & Easy & Mod. & Hard\\
				\hline
				
				\multirow{8}{*}{\rot{3D Perceptron}}
				& Point-RCNN$^{\dagger}$\cite{Shi2019b} & $\sim$10 & 86.96 & 75.64 & 70.70  & 92.13 & 87.39 & 82.72 & 74.96 & 58.82 & 52.53 & \textbf{82.56} & 67.24 & 60.28 \\
				&STD$^{\dagger}$\cite{yang2019std} & $<$20 & 87.95 & 79.71 & 75.09 & 94.74 & 89.19 & \textbf{86.42} & 78.69 & 61.59 & 55.30 & 81.36 & 67.23 & 59.35 \\
				&Fast Point-RCNN$^{\dagger}$\cite{chen2019fast} & $\sim$20 & 85.29 & 77.40 & 70.24 & 90.87 & 87.84 & 80.52 & - & - & - & - & - &- \\
				&PV-RCNN$^{\dagger}$\cite{shi2020pv} & $\sim$15 & 90.25 & 81.43 & 76.82 & 94.98 & 90.65 & 86.14 & 78.60 & 63.71 & \textbf{57.65} & 82.49 & \textbf{68.89} & \textbf{62.41} \\
				&SECOND\cite{yan2018second} & $\sim$30 & 83.34 & 72.55 & 65.82 & 89.39 & 83.77 & 78.59 & 71.33 & 52.08 & 45.83 & 76.50 & 56.05 & 49.45 \\
				&Voxel-RCNN$^{\dagger}$\cite{Deng2020} & $\sim$25 & \textbf{90.90} & \textbf{81.62} & \textbf{77.06} & - & - & - & - & - & - & - & - & - \\
				& SA-SSD\cite{He2020} & $\sim$25 & 88.75 & 79.79 & 74.16 & \textbf{95.03} & \textbf{91.03} & 85.96 & - & - & - & - & - & - \\
				& 3DSSD\cite{Yang2020} & $\sim$25 & 88.36 & 79.57 & 74.55 & - & - & - & \textbf{82.48} & \textbf{64.10} & 56.90 & - & - & - \\
				\hline
				\hline
				\multirow{3}{*}{\rot{BEV}}
				&PIXOR\cite{Yang2018} & $\sim$10 & - & - & - & 81.70 & 77.05 & 72.95 & - & - & - & - & - & - \\
				&HDNET\cite{yang2018hdnet} & $\sim$25 & - & - & - & 89.14 & 86.57 & 78.32 & - & - & - & - & - & - \\
				&PointPillars\cite{Lang2019} & $\sim$45 & 82.58 & 74.31 & 68.99 & 90.07 & 86.56 & 82.81 & 77.10 & 58.65 & 51.92 & 79.90 & 62.73 & 55.58 \\
				
				\hline
                &DV-Det (Ours)$^{\dagger}$ & \textbf{$\sim$75} & \textbf{85.32} & \textbf{76.74} & \textbf{70.02} & \textbf{91.53} & \textbf{87.65} & \textbf{83.27} & \textbf{77.94} & \textbf{58.77} & \textbf{52.49} & \textbf{80.92} & \textbf{63.08} & \textbf{57.32} \\
				\hline
				\multicolumn{15}{l}{$\dagger$ Two-stage model} \\
			\end{tabular}
		}
	\end{center}
\vspace{-0.2cm}
	\caption{Performance comparison on the KITTI testing dataset based on the mean Average Precision (mAP) with 40 recall positions.}
	\label{tab:kitti}
	\vspace{-0.2cm}
\end{table*} 

\subsection{Evaluation on Waymo Open Dataset}
Apart from the KITTI dataset, we also test our model on the Waymo Open Dataset to give a comprehensive evaluation. We train our model based on 798 training sequences with $\sim$158k point cloud scenes, and test our performance on 202 validation sequences with $\sim$40k point cloud scenes. Different from KITTI dataset, Waymo Open Dataset possesses a full 360\degree view angle with denser point cloud point cloud, which asks for more efficient models to achieve the real-time processing speed. We evaluate our method based on the official tools released by Waymo. The mAP performance of DV-Det is reported in Table \ref{tab:waymo} for vehicle detection with IoU threshold of 0.7, and we only consider the examples with difficulty LEVEL\_1 (objects possess more than 5 LiDAR points).

\begin{table*}[h!]
	\small 
	\vspace{-1mm}
    \begin{center}
        \scalebox{0.9}[0.9]{
            \setlength\tabcolsep{2pt}
            \begin{tabular}{c|c|cccc|cccc}
            \hline
            \multirow{2}{*}{Method} &
            Speed & 
            \multicolumn{4}{c|}{3D mAP (IoU=0.7)} & \multicolumn{4}{c}{BEV mAP (IoU=0.7)} \\
            &  (FPS) & Overall & 0-30m & 30-50m & 50m-Inf & Overall & 0-30m & 30-50m & 50m-Inf  \\
            \hline 
            PV-RCNN\cite{Shi2019b} & $<$5 & 70.30 & 91.92 & 69.21 & 42.17 & 82.96 & 97.53 & 82.99 & 64.97 \\
            Voxel-RCNN\cite{Deng2020} & $<$10 & 75.59 & 92.49 & 74.09 & 53.15 & 88.19 & 97.62 & 87.34 & 77.70 \\
            PointPillar\cite{Lang2019} & $\sim$15 & 56.62 & 81.01 & 51.75 & 27.94 & 75.57 & 92.10 & 74.06 & 55.47\\
            \hline 
            DV-Det (Ours) & \textbf{$\sim$25} & 63.42 & 85.72 & 55.89 & 30.73 & 78.38 & 95.12 & 76.93 & 59.36\\
            \hline 
            \end{tabular}
        }
    \end{center}
    \caption{Performance (mAP) comparison on the Waymo Open Dataset with 202 validation sequences for vehicle class.}
    \label{tab:waymo}
\end{table*}

Our DV-Det is overperformed by the state-of-the-art 3D perceptron based methods like Voxel-RCNN regarding the detection accuracy, but our method can run at a speed of 25 FPS, which is even faster than the 2D BEV methods like PointPillar with obviously higher accuracy. 

\section{Conclusion}

We present the DV-Det, a novel two-stage framework for efficient 3D object detection from point clouds. Our method combines the advantages from both point-wise methods and grid based methods, and as a pure 3D point cloud based method, DV-Det overperforms the 2D BEV based methods by a large margin regarding both efficiency and accuracy for the first time. With innovations like grid based point downsampling and dynamic voxelization techniques, we achieve satisfactory performance on KITTI and Waymo benchmark dataset. We hope our method can be inspiring and encourage more future works.

\pagebreak
\small
\bibliography{reference}

\begin{thebibliography}{10}

\bibitem{Chen2017}
Xiaozhi Chen, Huimin Ma, Ji~Wan, Bo~Li, and Tian Xia.
\newblock Multi-view 3d object detection network for autonomous driving.
\newblock 2017.

\bibitem{chen2019fast}
Yilun Chen, Shu Liu, Xiaoyong Shen, and Jiaya Jia.
\newblock Fast point r-cnn.
\newblock pages 9775--9784, 2019.

\bibitem{Deng2020}
Jiajun Deng, Shaoshuai Shi, Peiwei Li, Wengang Zhou, Yanyong Zhang, and
  Houqiang Li.
\newblock Voxel r-cnn: Towards high performance voxel-based 3d object
  detection.
\newblock {\em arXiv preprint arXiv:2012.15712}, 2020.

\bibitem{Geiger2012}
Andreas Geiger, Philip Lenz, and Raquel Urtasun.
\newblock Are we ready for autonomous driving? the kitti vision benchmark
  suite.
\newblock pages 3354--3361, 2012.

\bibitem{Graham2017}
Benjamin Graham and Laurens van~der Maaten.
\newblock Submanifold sparse convolutional networks.
\newblock 6 2017.

\bibitem{He2020}
Chenhang He, Hui Zeng, Jianqiang Huang, Xian-Sheng Hua, and Lei Zhang.
\newblock Structure aware single-stage 3d object detection from point cloud.
\newblock pages 11873--11882, 2020.

\bibitem{hu2020randla}
Qingyong Hu, Bo~Yang, Linhai Xie, Stefano Rosa, Yulan Guo, Zhihua Wang, Niki
  Trigoni, and Andrew Markham.
\newblock Randla-net: Efficient semantic segmentation of large-scale point
  clouds.
\newblock pages 11108--11117, 2020.

\bibitem{hua2018pointwise}
Binh-Son Hua, Minh-Khoi Tran, and Sai-Kit Yeung.
\newblock Pointwise convolutional neural networks.
\newblock In {\em Proceedings of the IEEE Conference on Computer Vision and
  Pattern Recognition}, pages 984--993, 2018.

\bibitem{Lang2019}
Alex~H. Lang, Sourabh Vora, Holger Caesar, Lubing Zhou, Jiong Yang, and Oscar
  Beijbom.
\newblock Pointpillars: Fast encoders for object detection from point clouds.
\newblock 2019.

\bibitem{Liang2019}
Ming Liang, Bin Yang, Yun Chen, Rui Hu, and Raquel Urtasun.
\newblock Multi-task multi-sensor fusion for 3d object detection.
\newblock 2019.

\bibitem{Liang2018deep}
Ming Liang, Bin Yang, Shenlong Wang, and Raquel Urtasun.
\newblock Deep continuous fusion for multi-sensor 3d object detection.
\newblock In {\em Proceedings of the European Conference on Computer Vision
  (ECCV)}, pages 641--656, 2018.

\bibitem{qi2019deep}
Charles~R Qi, Or~Litany, Kaiming He, and Leonidas~J Guibas.
\newblock Deep hough voting for 3d object detection in point clouds.
\newblock pages 9277--9286, 2019.

\bibitem{qi2018frustum}
Charles~R Qi, Wei Liu, Chenxia Wu, Hao Su, and Leonidas~J Guibas.
\newblock Frustum pointnets for 3d object detection from rgb-d data.
\newblock pages 918--927, 2018.

\bibitem{Qi2017}
Charles~R. Qi, Hao Su, Kaichun Mo, and Leonidas~J. Guibas.
\newblock Pointnet: Deep learning on point sets for 3d classification and
  segmentation.
\newblock volume 2017-Janua, pages 77--85, 2017.

\bibitem{Qi2017a}
Charles~R. Qi, Li~Yi, Hao Su, and Leonidas~J. Guibas.
\newblock Pointnet++: Deep hierarchical feature learning on point sets in a
  metric space.
\newblock 2017.

\bibitem{rezatofighi2019generalized}
Hamid Rezatofighi, Nathan Tsoi, JunYoung Gwak, Amir Sadeghian, Ian Reid, and
  Silvio Savarese.
\newblock Generalized intersection over union: A metric and a loss for bounding
  box regression.
\newblock In {\em Proceedings of the IEEE/CVF Conference on Computer Vision and
  Pattern Recognition}, pages 658--666, 2019.

\bibitem{Ronneberger2015}
Olaf Ronneberger, Philipp Fischer, and Thomas Brox.
\newblock U-net: Convolutional networks for biomedical image segmentation.
\newblock volume 9351, pages 234--241, 5 2015.

\bibitem{shi2020pv}
Shaoshuai Shi, Chaoxu Guo, Li~Jiang, Zhe Wang, Jianping Shi, Xiaogang Wang, and
  Hongsheng Li.
\newblock Pv-rcnn: Point-voxel feature set abstraction for 3d object detection.
\newblock pages 10529--10538, 2020.

\bibitem{Shi2019b}
Shaoshuai Shi, Xiaogang Wang, and Hongsheng Li.
\newblock Pointrcnn: 3d object proposal generation and detection from point
  cloud.
\newblock 2019.

\bibitem{Shi2019a}
Shaoshuai Shi, Zhe Wang, Jianping Shi, Xiaogang Wang, and Hongsheng Li.
\newblock Part-a2 net: 3d part-aware and aggregation neural network for object
  detection.
\newblock {\em Proceedings of the IEEE Computer Society Conference on Computer
  Vision and Pattern Recognition}, 2019.

\bibitem{Sun2020}
Pei Sun, Henrik Kretzschmar, Xerxes Dotiwalla, Aurelien Chouard, Vijaysai
  Patnaik, Paul Tsui, James Guo, Yin Zhou, Yuning Chai, Benjamin Caine, et~al.
\newblock Scalability in perception for autonomous driving: Waymo open dataset.
\newblock pages 2446--2454, 2020.

\bibitem{yan2018second}
Yan Yan, Yuxing Mao, and Bo~Li.
\newblock Second: Sparsely embedded convolutional detection.
\newblock {\em Sensors}, 18:3337, 2018.

\bibitem{yang2018hdnet}
Bin Yang, Ming Liang, and Raquel Urtasun.
\newblock Hdnet: Exploiting hd maps for 3d object detection.
\newblock In {\em Conference on Robot Learning}, pages 146--155. PMLR, 2018.

\bibitem{yang2018pixor}
Bin Yang, Wenjie Luo, and Raquel Urtasun.
\newblock Pixor: Real-time 3d object detection from point clouds.
\newblock In {\em Proceedings of the IEEE conference on Computer Vision and
  Pattern Recognition}, pages 7652--7660, 2018.

\bibitem{Yang2018}
Bin Yang, Wenjie Luo, and Raquel Urtasun.
\newblock Pixor: Real-time 3d object detection from point clouds.
\newblock 2018.

\bibitem{Yang2020}
Zetong Yang, Yanan Sun, Shu Liu, and Jiaya Jia.
\newblock 3dssd: Point-based 3d single stage object detector.
\newblock pages 11040--11048, 2020.

\bibitem{yang2019std}
Zetong Yang, Yanan Sun, Shu Liu, Xiaoyong Shen, and Jiaya Jia.
\newblock Std: Sparse-to-dense 3d object detector for point cloud.
\newblock pages 1951--1960, 2019.

\bibitem{zhou2019iou}
Dingfu Zhou, Jin Fang, Xibin Song, Chenye Guan, Junbo Yin, Yuchao Dai, and
  Ruigang Yang.
\newblock Iou loss for 2d/3d object detection.
\newblock pages 85--94, 2019.

\bibitem{zhou2018voxelnet}
Yin Zhou and Oncel Tuzel.
\newblock Voxelnet: End-to-end learning for point cloud based 3d object
  detection.
\newblock pages 4490--4499, 2018.

\end{thebibliography}

\end{document}